\documentclass[10pt,twocolumn,letterpaper]{article}

\usepackage{cvpr}              

\definecolor{cvprblue}{rgb}{0.21,0.49,0.74}
\usepackage[pagebackref,breaklinks,colorlinks,allcolors=cvprblue]{hyperref}

\usepackage{amsmath}
\usepackage{amsfonts}
\usepackage{booktabs}
\usepackage{graphicx}
\usepackage{subcaption}
\usepackage{multirow}
\usepackage{bm}
\usepackage{graphicx}
\usepackage{booktabs}
\usepackage{multirow}
\usepackage{array}
\usepackage{algpseudocode}
\usepackage{listings}
\usepackage{subcaption}
\usepackage{subcaption}
\usepackage{caption}
\usepackage{latexsym}
\usepackage{xcolor}
\usepackage{tikz}

\def\paperID{7323} 
\def\confName{CVPR}
\def\confYear{2026}

\title{Where Does Vision Meet Language? Understanding and Refining Visual Fusion in MLLMs via Contrastive Attention}

\author{ 
    Shezheng Song$^1$, Shasha Li$^{2}$\thanks{Corresponding author.}, Shan Zhao$^{1*}$, Xiaopeng Li$^2$, Qian Wan$^3$, \\
    Chengyu Wang$^4$, Tianwei Yan$^5$, Jun Ma$^2$, Jie Yu$^2$.\\
    $^1$ Hefei University of Technology, 
    $^2$ National University of Defense Technology, \\
    $^3$ Central China Normal University, 
    $^4$ Hunan University, 
    $^5$ Chongqing Jiaotong University.\\
    {\tt\small betterszsong@gmail.com}
}

\begin{document}
\maketitle

\begin{abstract}
Multimodal Large Language Models (MLLMs) have achieved remarkable progress in vision–language understanding, yet how they internally integrate visual and textual information remains poorly understood. To bridge this gap, we perform a systematic layer-wise masking analysis across multiple architectures, revealing how visual–text fusion evolves within MLLMs. The results show that fusion emerges at several specific layers rather than being uniformly distributed across the network, and certain models exhibit a late-stage “review” phenomenon where visual signals are reactivated before output generation. Besides, we further analyze layer-wise attention evolution and observe persistent high-attention noise on irrelevant regions, along with gradually increasing attention on text-aligned areas. Guided by these insights, we introduce a training-free contrastive attention framework that models the transformation between early fusion and final layers to highlight meaningful attention shifts. Extensive experiments across various MLLMs and benchmarks validate our analysis and demonstrate that the proposed approach improves multimodal reasoning performance.
\end{abstract}

\section{Introduction}
 

Multimodal Large Language Models (MLLMs) \cite{liu2023llava, Qwen-VL} have achieved remarkable performance across a wide range of vision–language tasks \cite{goyal2017vqav2, marino2019okvqa}. Recent studies have explored the internal mechanisms of MLLMs, including the role of FFN and memory \cite{aflalo2022vl, chefer2021generic, lyu2022dime, stan2024lvlm}, the evolution and localization of visual features \cite{chefer2021generic, lyu2022dime, stan2024lvlm}, and the reduction of redundant visual tokens \cite{palit2023towards, zhao2024first}.
Besides, \citet{zhang2025cross} conduct a preliminary exploration of the contribution of visual information to final answers on the GQA dataset, while the \citet{palit2023towards} identify causal relationships between image patches and predictions through causal tracing.
Despite these advances, it is still unclear how visual information is progressively integrated across different layers of MLLMs. A deeper understanding of this process is essential for improving the interpretability, reliability, and performance.

%
To address this gap, we focus on the internal process of visual-textual information fusion: \textit{At which layers is visual information integrated, and how does integration in each layer affect task performance?}
To explore this, we conduct a comprehensive series of layer-wise masking experiments across seven representative MLLMs and six multimodal benchmarks, systematically evaluating the role of each layer in cross-modal fusion. Specifically, we apply masking to visual feature at different layers and observe the resulting changes in performance. A significant performance drop after masking indicates that the corresponding layer plays a critical role in visual integration, whereas a minimal impact suggests a weaker dependence on visual information. 
The results consistently indicate that multimodal \textbf{fusion emerges at several specific layers}, rather than being uniformly distributed across the network.
At these layers, removing visual input almost collapses task performance, while deeper layers become increasingly text-dominant. Interestingly, certain models such as LLaVA-1.5, LLaVA-Next, and InstructBLIP exhibit a “\textbf{review}” phenomenon at later stages, where visual information is reactivated before the final output. These findings offer an interpretable view of how vision–language fusion is distributed and temporally organized within MLLMs. Then an attention distance analysis supports these findings, showing that attention patterns stabilize after the fusion stage. Besides, we conduct an early output experiment, where responses are generated from intermediate layers rather than the final one. The results show that even outputs from mid-level layers can yield partially correct answers, suggesting that a substantial portion of visual–text fusion has already been achieved at these stages.

\begin{figure}[htb]
    \centering
    \includegraphics[width=\linewidth]{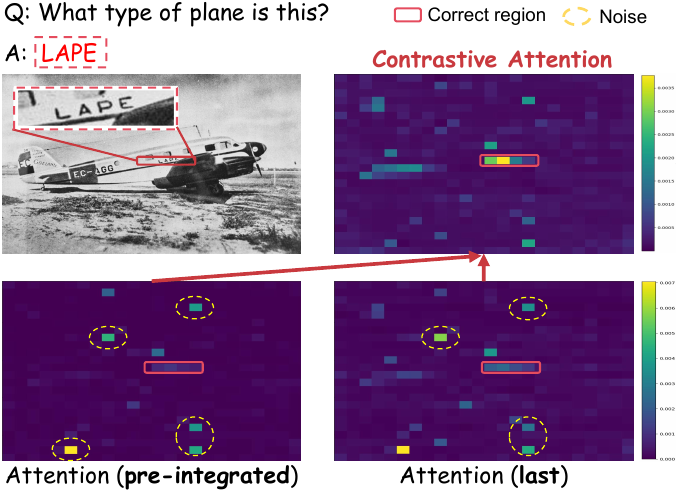}
    \caption{Contrastive attention (pre-integrated $\rightarrow$ last layer) showing focus shift and persistent high-attention noise.}
    \label{fig:intro}
\end{figure}
We further visualize the layer-wise evolution of attention to explore how models internally attend to visual cues. As shown in \Cref{fig:intro}, we find that the attention map of the final layer often exhibits substantial noise, frequently assigning high weights to visually irrelevant regions regardless of the input question. As illustrated by the yellow dashed circles in the figure, these \textbf{high-attention noise regions} consistently receive strong activation even though they bear no semantic relation to the queried object. In contrast, the truly relevant region, highlighted by the red bounding box, is often weakly attended or partially overlooked. This indicates that the model’s visual understanding is distracted. More importantly, this phenomenon is not limited to the final layer, since similar noisy regions persist across multiple earlier layers. This suggests that attention biases emerge early and propagate through the network rather than being reformed at each layer. The layer-to-layer persistence of noisy activation exposes a systematic attention bias.


To mitigate this systematic attention bias, we propose a contrastive attention mechanism that models how visual attention evolves across layers rather than relying on a single static attention map. Specifically, it captures the transformation from a \textit{pre-integrated layer}, which represents the stage where visual and textual features begin to interact but multimodal semantics are not yet fully fused, to the \textit{final layer}, which reflects the completed multimodal understanding. The pre-integrated layer serves as a reference point that reflects the model’s early fusion state, allowing us to contrast it against the final layer to capture the meaningful attention shifts induced by textual guidance.
As MLLMs process multimodal inputs under textual guidance, their visual attention gradually shifts from noisy or diffuse patterns toward task-relevant regions. By contrasting the attention between the pre-integrated and final layers, our method captures these meaningful text-guided attention transitions, reducing redundant visual noise and enhancing focus on semantically aligned areas.
We then leverage this contrastive attention to segment the original image, preserving the regions with strong attention change and re-inputting them into the model. This process reduces redundant visual noise and strengthens the model’s focus on semantically relevant content.
Unlike static approaches such as ViCrop \cite{zhang2025mllms}, which fix extraction layers per model, our method dynamically adapts to model- and dataset-specific fusion dynamics, enabling more general and interpretable multimodal alignment.
Our contributions are as follows:
\begin{itemize}
    \item We perform a comprehensive layer-wise investigation of how visual information is progressively fused in MLLMs. Through systematic visual masking, we identify key fusion layers and reveal a consistent shallow-to-middle fusion pattern, as well as a “review” phenomenon.
    \item We propose a training-free method based on contrastive attention, which captures the evolution of visual attention between a pre-integrated and the final layer. This mechanism effectively suppresses persistent high-attention noise and enhances text-guided visual focus.
    \item Extensive experiments across multiple MLLMs and multimodal benchmarks demonstrate that our method consistently improves performance and achieves state-of-the-art results without additional training.

\end{itemize}

\begin{figure*}[htb]
    \centering
    \includegraphics[width=\linewidth]{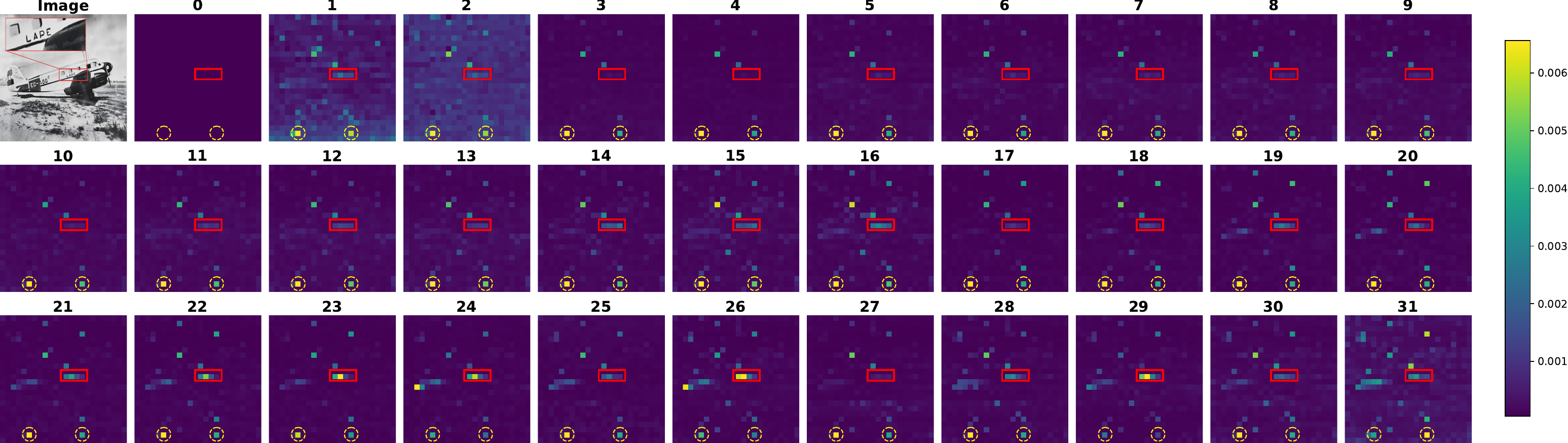}
    \caption{
    Layer-wise evolution of visual attention. The question is “What type of plane is this?”, where the key is “LAPE” logo, as indicated by the red box.
    The \textcolor{red}{red box} marks the correct target region,
    while the \textcolor{yellow!80!black}{yellow circles} highlight persistent high-attention noise.
    }
    \label{fig:all_attention}
\end{figure*}

\section{Related Work}
Early efforts to understand how information flows within language models primarily focused on transformers in unimodal settings. For example, studies like \cite{cao2020behind, frank2021vision} analyzed how attention mechanisms propagate syntactic and semantic information in transformers, while others \cite{aflalo2022vl, chefer2021generic, lyu2022dime, stan2024lvlm} explored the roles of feed-forward networks and memory representation in information transformation. These works provided foundational insights into token-level dependency and hierarchical abstraction but largely focused on language-only settings.
Recent studies examined how LLMs acquire knowledge across layers. \citet{lin2025multi} investigates the impact of visual fusion positions on MLLM performance, comparing external and internal integration strategies for incorporating visual information. \citet{jin2024exploring} introduced the concept of ``depth", and \citet{ju2024large} showed context is unevenly distributed across layers. 

A growing body of work has recently turned to the interpretability of MLLMs. For example, \citet{palit2023towards} extended causal tracing methods to MLLMs and showed that visual information progressively influences generation in later layers. \citet{zhao2024first} surveyed various interpretability techniques for MLLMs, categorizing insights at input, token, and attention levels. However, most of these efforts focus on saliency or input-output attribution and rarely dissect the internal fusion mechanism of visual features across layers.
Despite these efforts, what remains largely missing is a comprehensive understanding of how visual information is gradually injected and transformed across different layers within MLLMs. Existing works often rely on single-layer attention analysis or fixed fusion strategies, overlooking the progressive, multi-stage nature of visual integration.
A comprehensive understanding of how visual information is progressively injected in MLLMs is still lacking.

\begin{figure}
    \centering
    \includegraphics[width=0.75\linewidth]{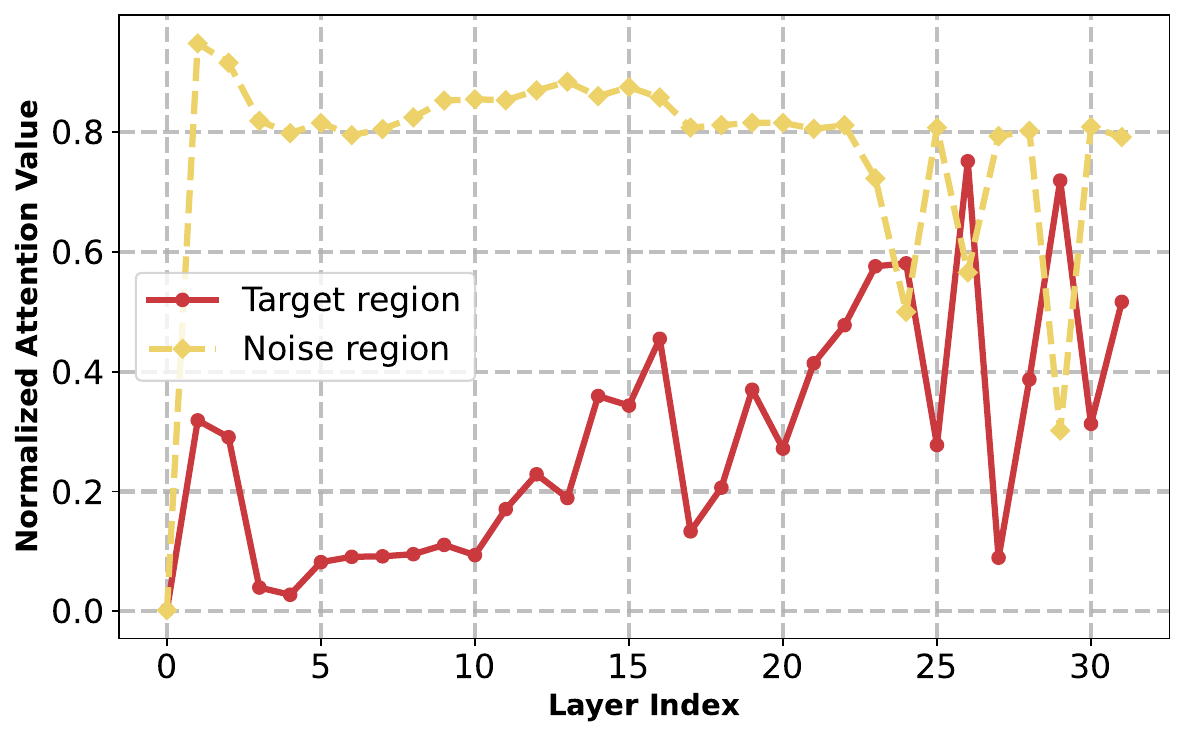}
    \caption{Layer-wise quantitative attention evolution.
    The red curve shows a gradual increase in target-region attention with depth, while the yellow curve remains high over irrelevant areas.}
    \label{fig:attn_evolution}
\end{figure}

\section{Layer-Wise Exploration of Visual Fusion}
This section investigates how visual and textual information are progressively fused across layers in MLLMs, providing the foundation for identifying fusion layers.

\subsection{Evolution of Layer-Wise Attention Patterns}
To illustrate the internal behavior of MLLMs, we visualize the layer-wise attention of LLaVA-1.5 for a representative example (\Cref{fig:all_attention}). The question is “\textit{What type of plane is this?},” where the key to answering is the “\texttt{LAPE}” logo on the aircraft.
In \Cref{fig:all_attention}, the attention evolves across layers. At first layer, attention is nearly uniform, indicating no meaningful focus. From the early layers, LLaVA begins forming a text-guided visual focus, with localized high-attention regions appearing around the “\texttt{LAPE}” area. As depth increases, attention on the correct region (red box) strengthens and peaks near layer 20, showing effective alignment with the textual cue.
However, we observe \textbf{persistent noisy activations} over irrelevant areas, such as the tail section (\textit{yellow dashed circles}). These noise regions remain strongly activated across all layers, revealing that the model retains and propagates early attention biases. 
While deeper layers (e.g., layer 29) can produce more accurate grounding, their effects differ across models and datasets. Approaches such as ViCrop \cite{zhang2025mllms} exploit this by manually selecting layers (e.g., layer 14 in InstructBLIP or layer 17 in LLaVA) for cleaner visual attention. Yet, this fixed-layer design demands model- or dataset-specific tuning, which reduces generality and scalability. This motivates our exploration of a dynamic mechanism.

We further conduct a quantitative analysis to verify that the trend is consistent across samples rather than a single case. Specifically, we randomly select 1,000 samples from the TextVQA dataset and manually annotate key regions (red boxes) and high-attention noise regions (yellow boxes). We then compute the average normalized attention of these regions across all 32 layers on LLaVA-1.5. As shown in Figure \ref{fig:attn_evolution}, the red curve shows a gradual increase in attention with depth, indicating improved focus on target regions, while the yellow curve stays consistently high, reflecting persistent noise on distractor areas.


To reduce the effect of persistent high-attention noise regions, we contrast attention distributions across layers instead of analyzing them individually. Since such noise often propagates through layers, single-layer attention tends to inherit and amplify it. By comparing attention distributions, we can suppress these shared noise patterns and highlight layer-specific shifts that reflect text-guided visual evolution.
The key is selecting layers that capture the transition where visual understanding emerges, as shallow layers remain unaligned while deeper layers provide limited contrast.

\subsection{Visual Masking Experiments}

\begin{figure}[htb]
    \centering
    \includegraphics[width=\linewidth]{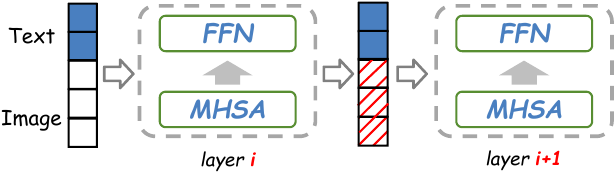}
    \caption{Illustration of layer-wise visual masking. Red regions indicate the masked image tokens at a specific layer.}
    \label{fig:mask_exp_setting}
\end{figure}

\begin{figure*}[htb]
    \centering
    \begin{subfigure}[b]{0.48\linewidth}
        \centering
        \includegraphics[width=\linewidth]{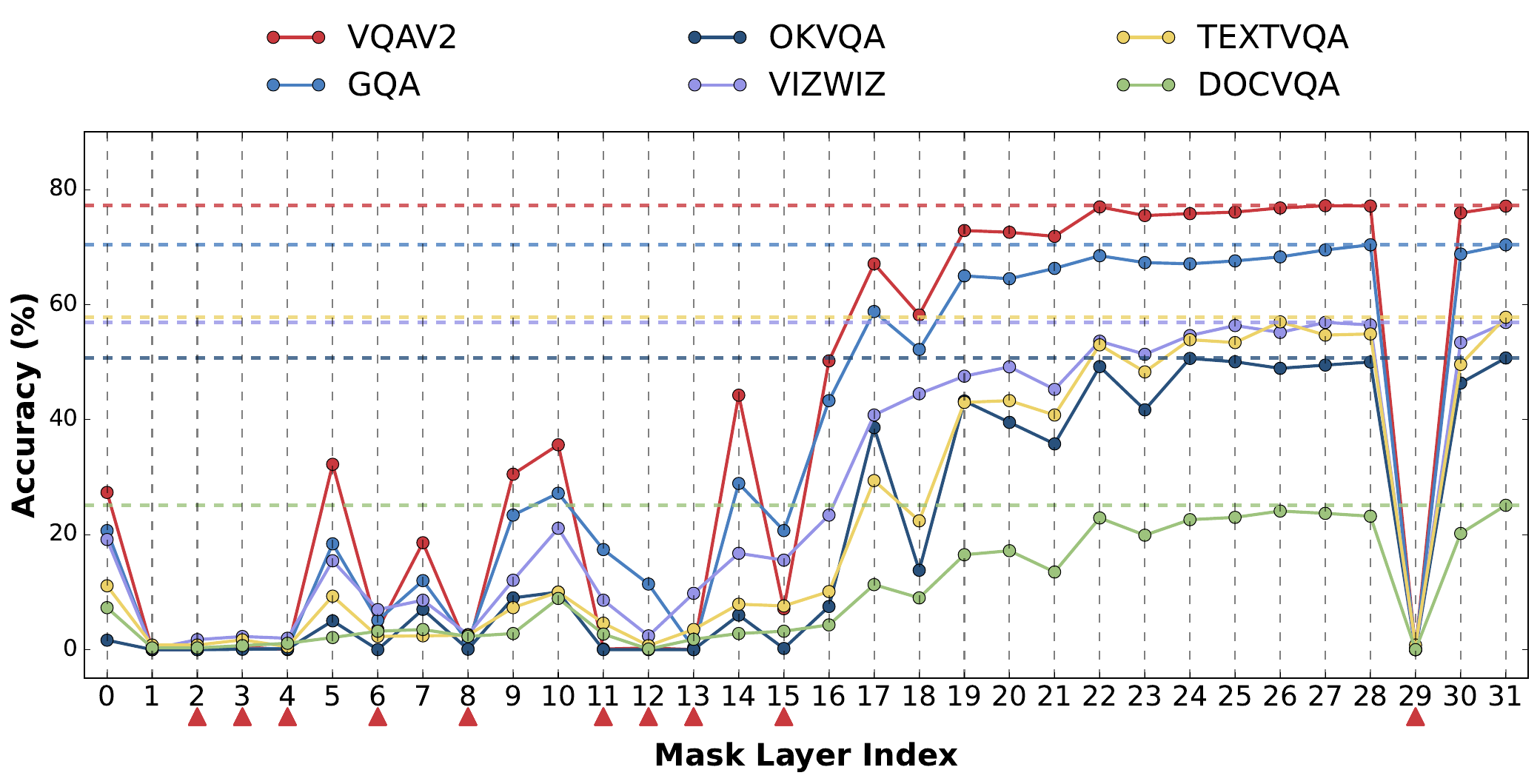}
        \caption{LLaVA-1.5 \cite{liu2023llava}}
        \label{fig:mask_llava15}
    \end{subfigure}
    \begin{subfigure}[b]{0.48\linewidth}
        \centering
        \includegraphics[width=\linewidth]{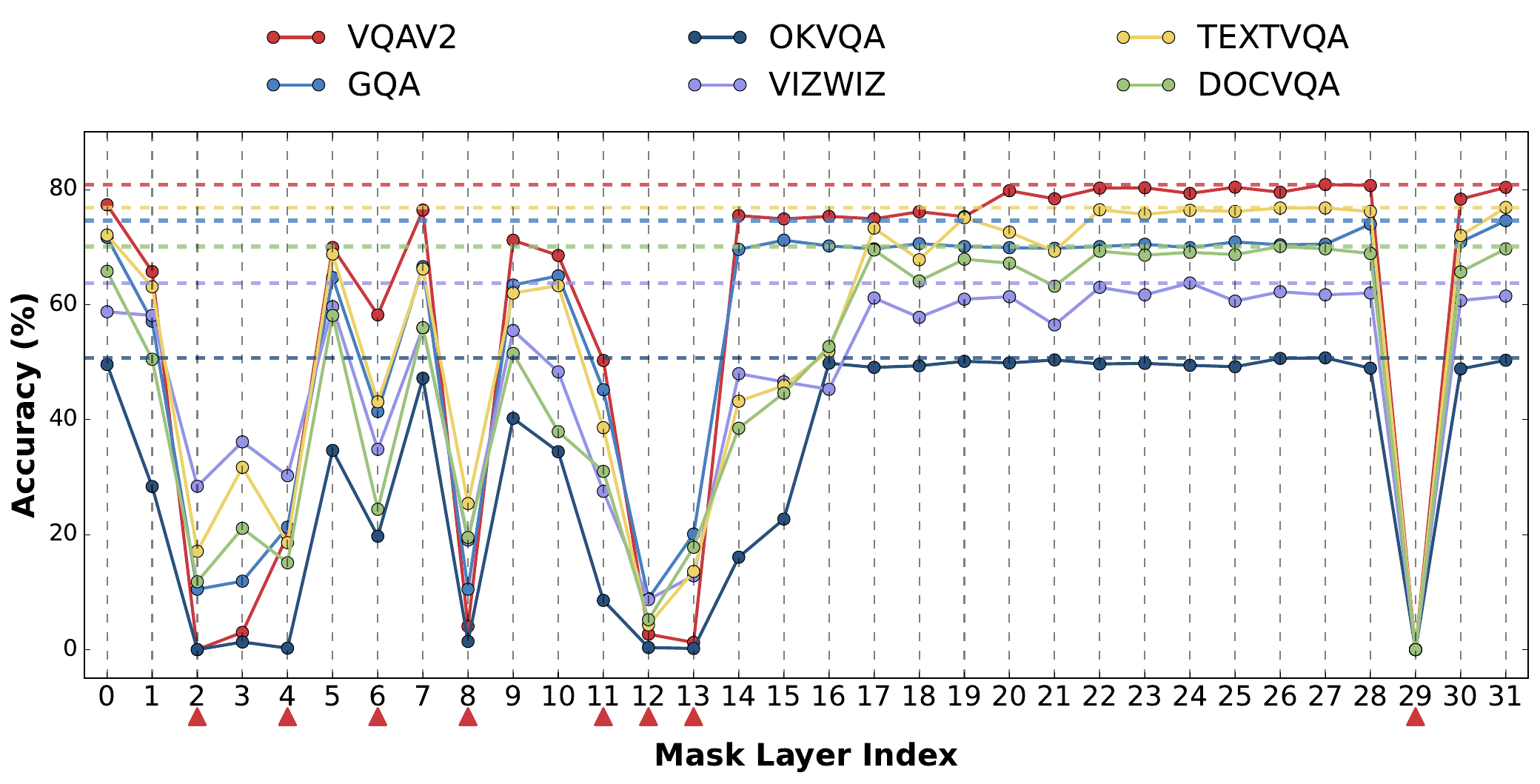}
        \caption{LLaVA-Next \cite{liu2023improvedllava}}
        \label{fig:mask_llavaNext}
    \end{subfigure}
    \\
    \begin{subfigure}[b]{0.48\linewidth}
        \centering
        \includegraphics[width=\linewidth]{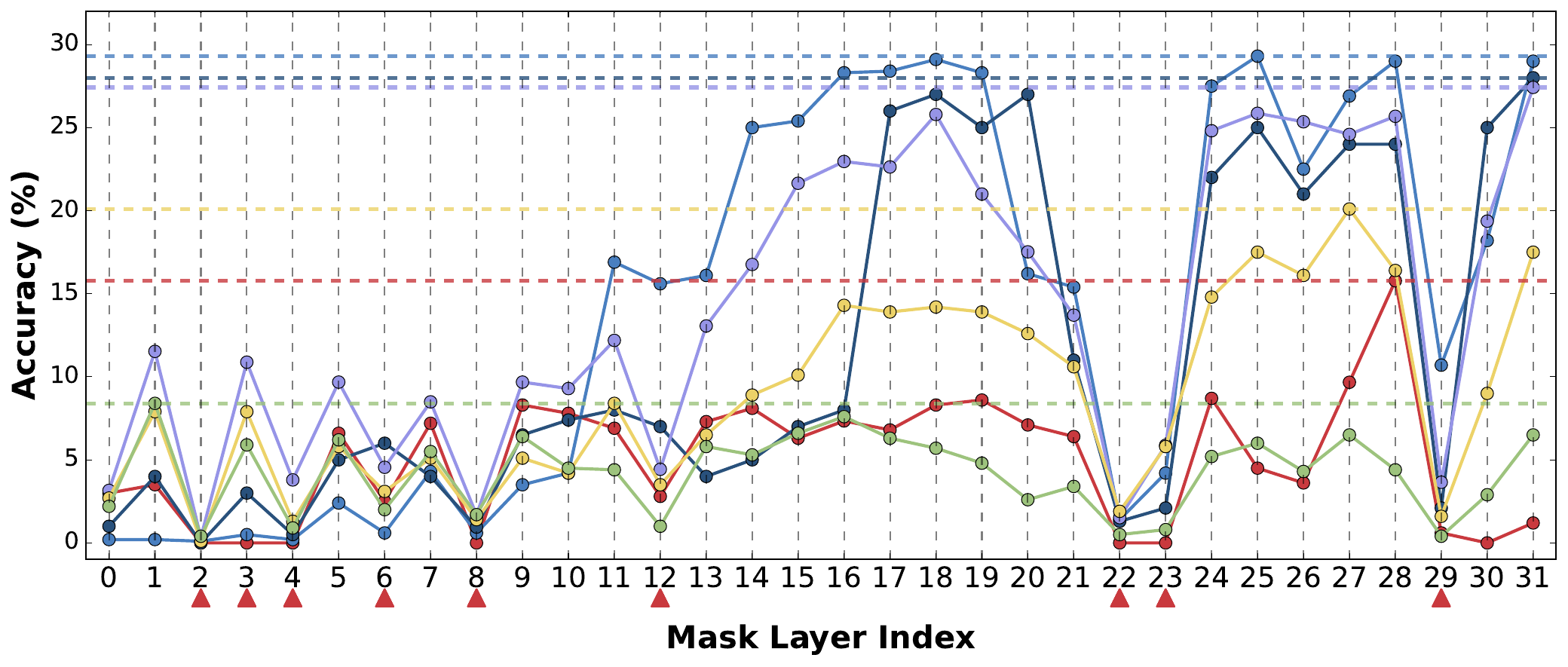}
        \caption{InstructBLIP \cite{instructblip}}
        \label{fig:mask_insblip}
    \end{subfigure}
    \begin{subfigure}[b]{0.48\linewidth}
        \centering
        \includegraphics[width=\linewidth]{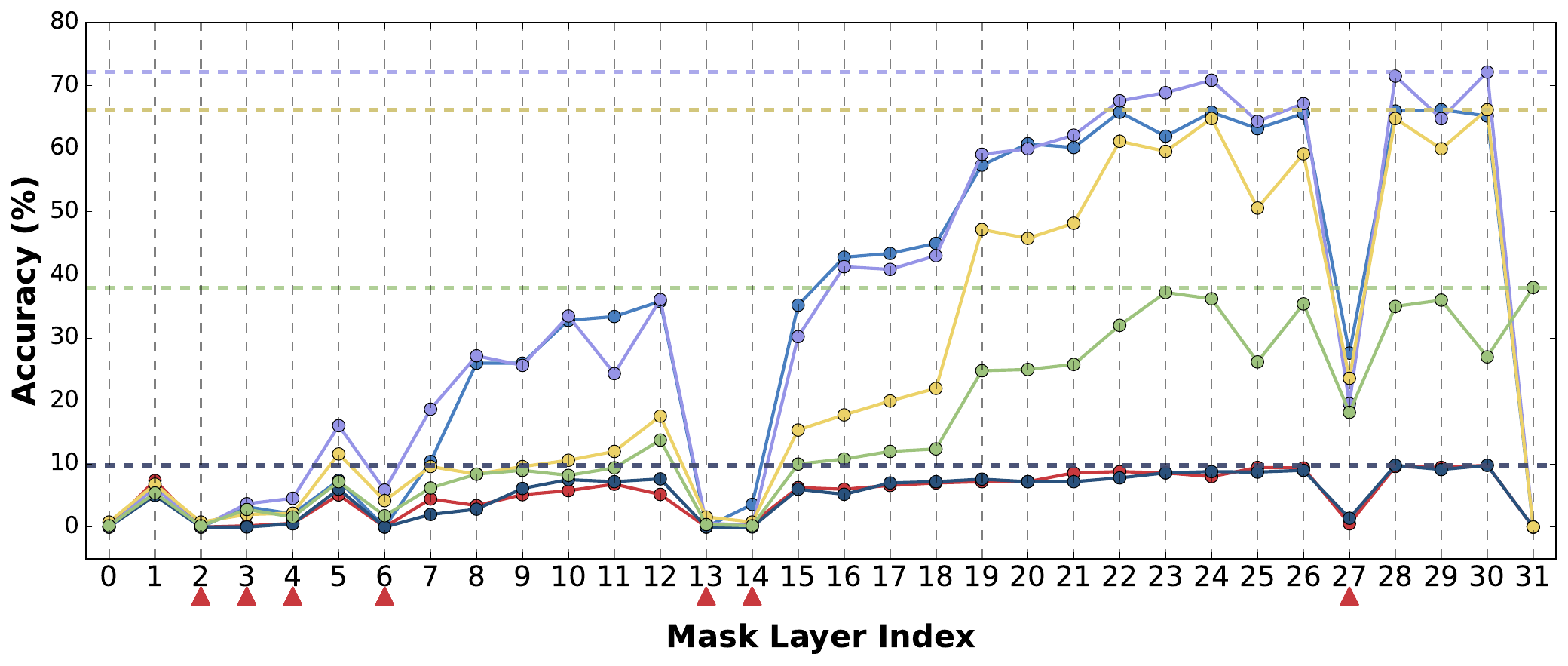}
        \caption{VIPLLaVA \cite{cai2024vipllava}}
        \label{fig:mask_VipLLaVA}
    \end{subfigure}
    \caption{
        Layer-wise masking performance with ``Review" at deep layer (e.g. 29). The horizontal dashed lines represent the original accuracies without masking, while the solid points indicate the actual accuracies after masking the image tokens at the corresponding layer.
    }
    \label{fig:mask_exp_llava_series}
\end{figure*}


To identify this transition point, as shown in \Cref{fig:mask_exp_setting}, we conduct a layer-wise masking exploration.
In MLLMs, the text and image representations are processed together through sequential transformer layers, where each layer passes its hidden states to the next while keeping the same feature dimension. The image-related tokens are located at fixed positions in this sequence, allowing us to precisely target them at any layer.
During the masking exploration, we set the features corresponding to image tokens to zero at specific layers to remove visual information. By examining the effect of the intervention on model responses, we identify layers that depend on visual information. Performance variations across datasets reveal each layer’s reliance on image features, where \textbf{a notable drop after masking indicates strong visual dependence}.
Experiments are taken on multimodal datasets: VQAv2 \cite{goyal2017vqav2}, GQA \cite{hudson2019gqa}, TextVQA \cite{singh2019textvqa}, OKVQA \cite{marino2019okvqa}, VizWiz \cite{gurari2018vizwiz}, and DocVQA \cite{mathew2021docvqa}.

\subsection{Layer-Wise Fusion Behavior}
\label{sec:Layer-Wise Fusion Behavior}

As shown in \Cref{fig:mask_exp_llava_series} and \Cref{fig:mask_exp_qwen_series}, layer-wise visual masking results on seven representative MLLMs reveal \textbf{consistent performance trends across models and datasets}. The magnitude of accuracy drop reflects the extent to which each layer relies on visual input for fusion.

\subsubsection{Image-Text Fusion Stage}
\label{sec:Image-Text Fusion Stage}
\begin{figure}[htbp]
    \centering
    \begin{subfigure}[h]{\linewidth}
        \centering
        \includegraphics[width=\linewidth]{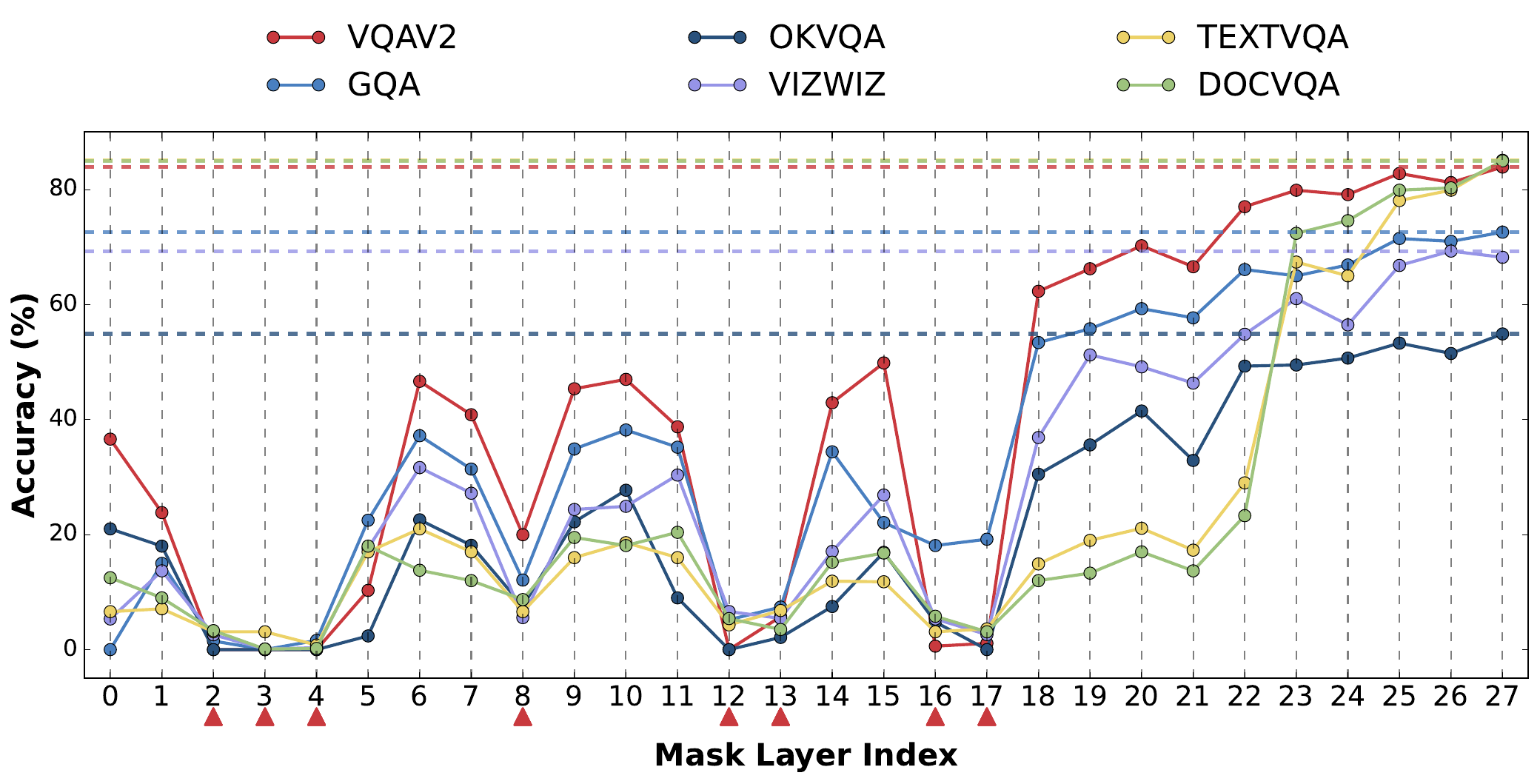}
        \caption{Qwen2VL \cite{wang2024qwen2}}
        \label{fig:mask_qwen2vl}
    \end{subfigure}
    \\
    \begin{subfigure}[h]{\linewidth}
        \centering
        \includegraphics[width=\linewidth]{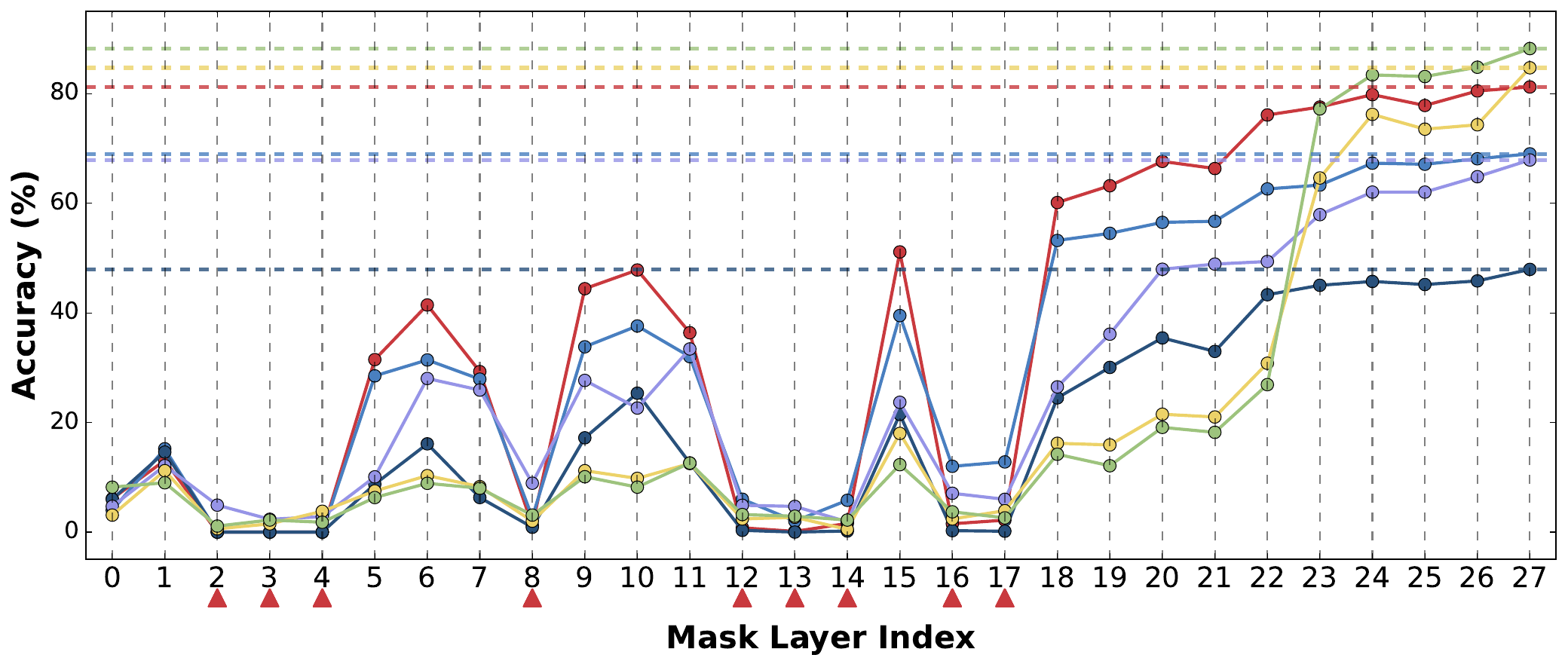}
        \caption{Qwen2.5VL \cite{qwen2.5-VL}}
        \label{fig:mask_qwen25vl}
    \end{subfigure}
    \\
    \begin{subfigure}[h]{\linewidth}
        \centering
        \includegraphics[width=\linewidth]{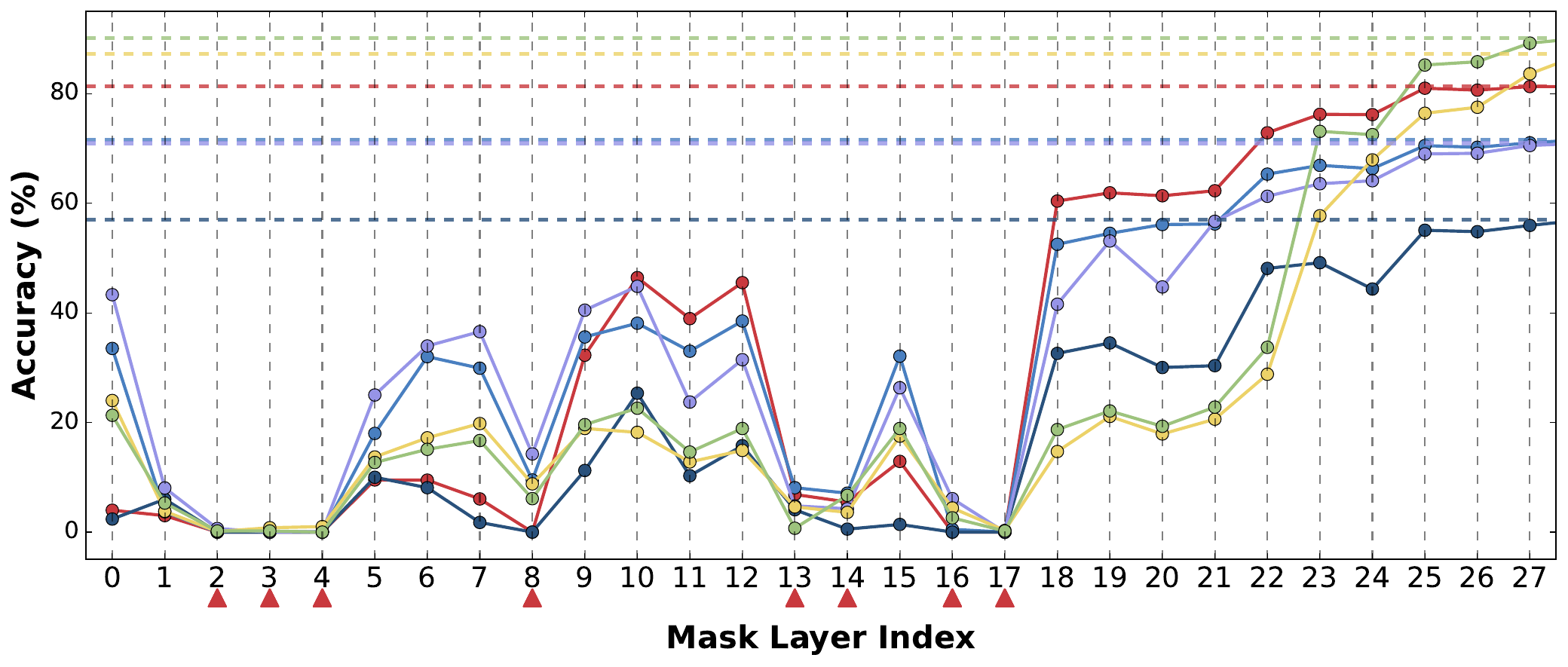}
        \caption{LLaVA-Onevision \cite{li2024llavaonevision}}
        \label{fig:mask_llavaOV}
    \end{subfigure}

    \caption{
        Layer-wise masking on MLLMs without ``Review". The dashed lines denote the original accuracies, and the solid points show the results after masking image tokens at each layer.
    }
    \label{fig:mask_exp_qwen_series}
\end{figure}

    The results in \Cref{fig:mask_exp_llava_series} and \Cref{fig:mask_exp_qwen_series} consistently reveal a two-stage fusion pattern across all models and datasets.
    At the \textbf{shallow} layers, masking visual tokens causes a sharp performance decline, indicating that image features have not yet been fully integrated into the multimodal representations. The model heavily depends on visual input at these stages, and removing it deprives the system of critical perceptual cues necessary for accurate reasoning.
    As the layers \textbf{deepen}, this dependence gradually weakens. Beyond layer 18–20, performance stabilizes even when visual input is masked, suggesting that multimodal fusion is largely complete and the model shifts to text-dominant reasoning based on encoded semantics.

\begin{figure}[htb]
    \centering
    \includegraphics[width=\linewidth]{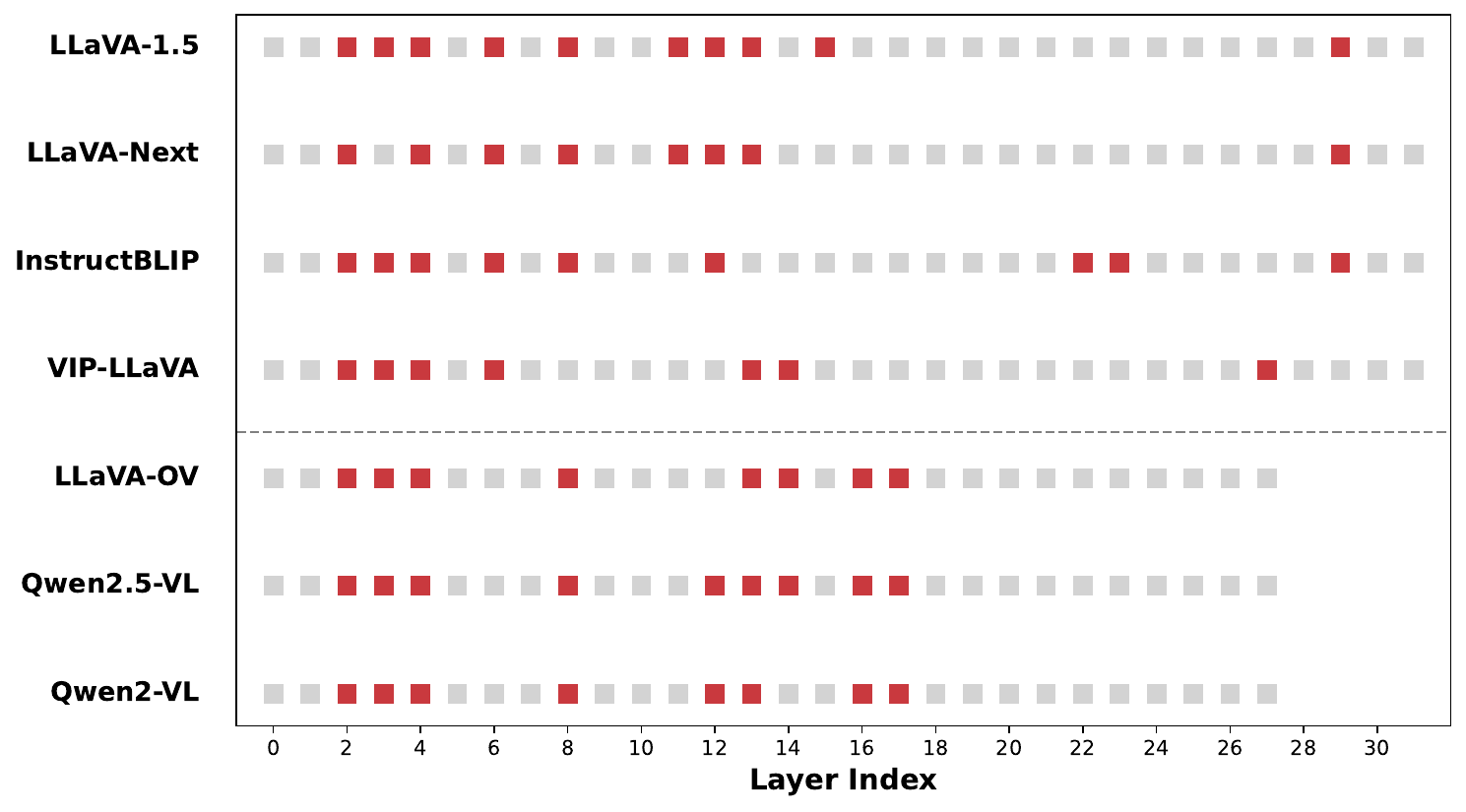}
    \caption{Distribution of Fusion Layers across MLLMs.
    Each red square marks a layer identified as a key fusion point in the masking experiments from \Cref{fig:mask_exp_llava_series} and \Cref{fig:mask_exp_qwen_series}.}
    \label{fig:highlight}
\end{figure}

\subsubsection{Shallow Fusion Layers}
\label{sec:Shallow Fusion Layers}
We find that \textbf{masking visual information at certain specific layers leads to a dramatic drop} in model performance. 
In \Cref{fig:mask_exp_llava_series} and \Cref{fig:mask_exp_qwen_series}, these fusion layers are highlighted with red triangles.
To further quantify this phenomenon, \Cref{fig:highlight} summarizes the distribution of fusion layers across seven representative MLLMs. Despite differences in architectures and training strategies, their fusion layers show a highly similar distribution pattern. 
This consistency indicates that multimodal fusion does not occur uniformly across all layers. Instead, it becomes particularly strong at a few key layers that play a dominant role in cross-modal integration. Although the exact positions of these layers may vary slightly among different models, their overall distribution remains remarkably stable.

\subsubsection{Review Phenomenon at Deep Layer}
\label{sec:Review Phenomenon at Deep Layer}
After the shallow fusion layers, model performance stabilizes and becomes largely insensitive to visual masking, indicating that the multimodal fusion process has been largely completed. As illustrated in \Cref{fig:highlight}, the fusion behavior primarily occurs within the first twenty layers.
However, in several models such as LLaVA-1.5, LLaVA-Next, InstructBLIP, and VIP-LLaVA, we observe an unusual phenomenon: at a deeper layer, masking visual features once again leads to a noticeable accuracy drop, suggesting that the model re-engages in a secondary fusion process.
We refer to this layer as a review layer, where the model appears to “\textbf{revisit}” visual information after the initial fusion has been completed. This mirrors human behavior, where one might \textbf{glance back at image} before making a decision.

\subsubsection{Hellinger Distance Analysis of Layer Attention}

\begin{figure}[htb]
    \centering
    \includegraphics[width=\linewidth]{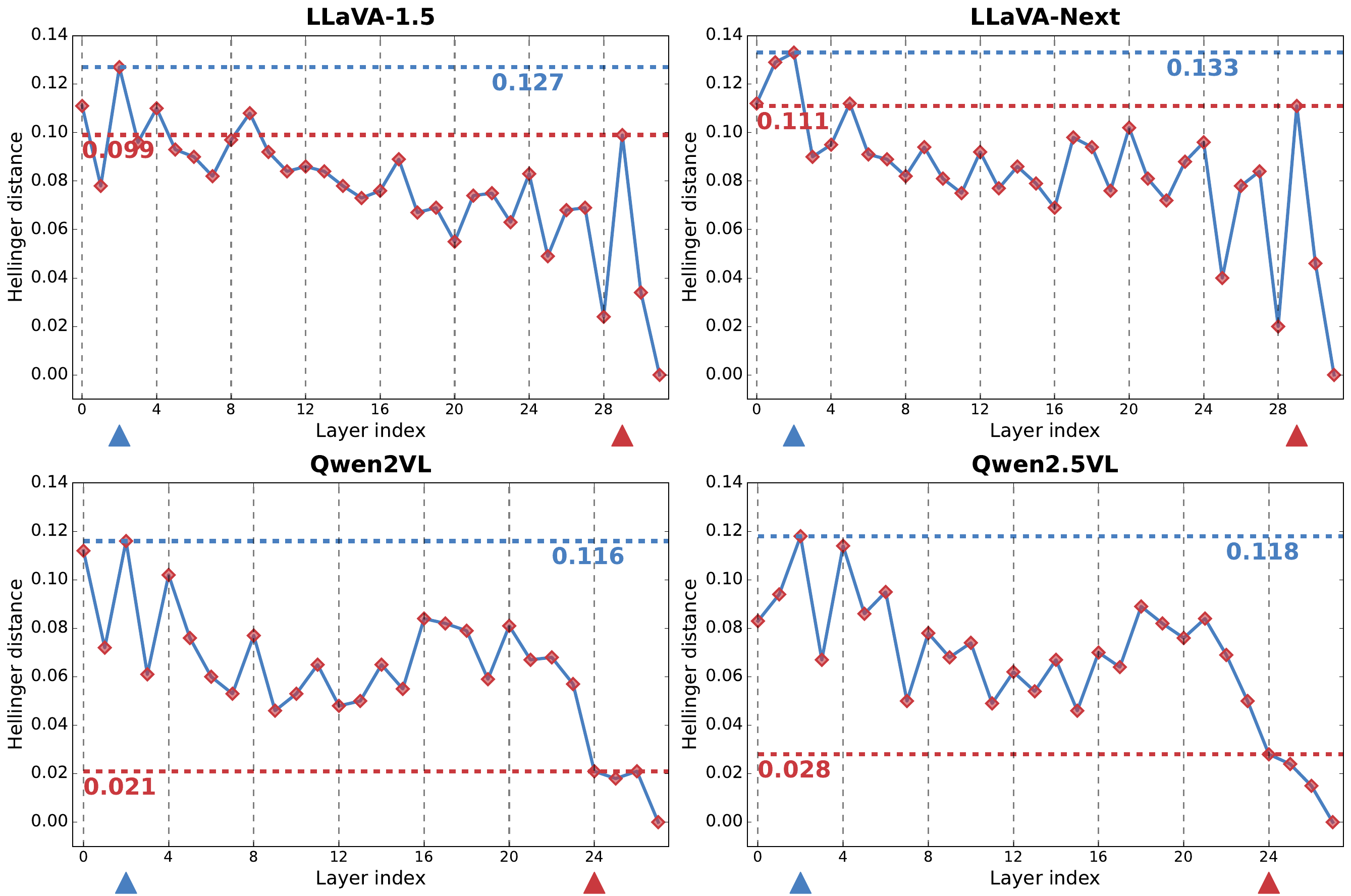}
    \caption{Hellinger distance between each layer and the final layer. Red triangles mark layer 29 in LLaVA models, where the “review” behavior reappears, and layer 24 in Qwen models, where the fusion process becomes stable. Blue dashed lines indicate the maximum distance.}
    \label{fig:integrated_layer}
\end{figure}

To further interpret the above phenomena, we compute the Hellinger distance between the attention map of each layer and that of the final layer, as shown in \Cref{fig:integrated_layer}. This metric quantifies how much each layer attention distribution diverges from the final visual focus. A larger distance indicates that the layer attention pattern remains less fused and more distinct from the integrated representation.

Across models, shallow layers show relatively large distances, which gradually decrease with depth. This consistent decreasing trend suggests that visual and textual information are progressively integrated along the model hierarchy, and the fusion process is largely completed within the early-to-middle layers (approximately before layer 20). This observation aligns with the layer-wise masking results in \Cref{sec:Image-Text Fusion Stage}, where masking in early layers causes substantial performance degradation, indicating that these layers serve as the primary sites of visual–text fusion.

However, in LLaVA-1.5 and LLaVA-Next, the Hellinger distance rises again at layer 29 after remaining low for several layers. This renewed divergence from the final representation corresponds to the “review” phenomenon discussed in \Cref{sec:Review Phenomenon at Deep Layer}, where visual information is reactivated at a late stage. In contrast, Qwen2VL and Qwen2.5VL maintain small and stable distances after layer 24, showing no sign of such late-stage re-fusion.

\subsection{When Does Fusion Complete? Early Output Analysis}

\begin{figure}[htb]
    \centering
    \includegraphics[width=\linewidth]{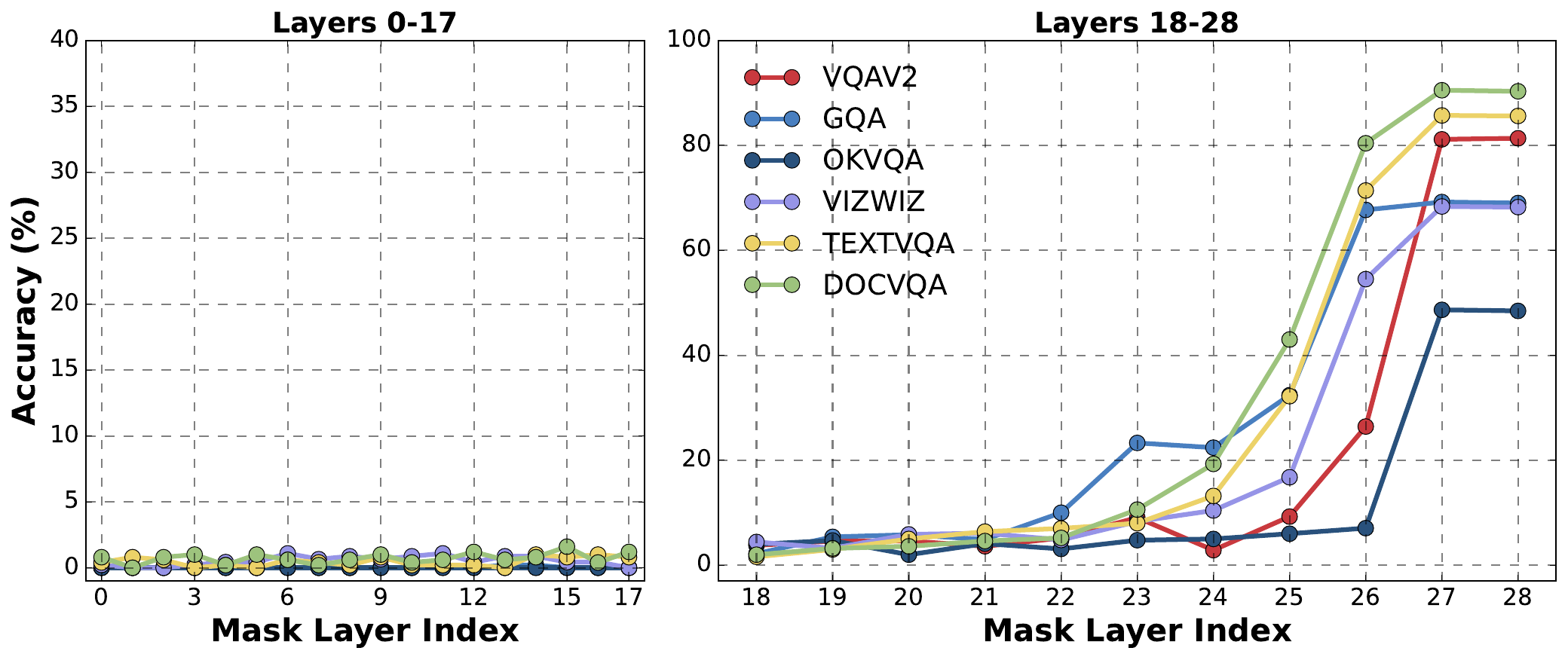}
    \caption{Layerwise output performance of the Qwen2.5VL on VQA. Accuracy rises notably at deep layers, suggesting that image–text fusion is completed at this stage.}
    \label{fig:pre_output}
\end{figure}

The experiment in \Cref{fig:pre_output} investigates the performance of the Qwen2.5VL when the output is generated early, aiming to examine how image–text integration evolves. Specifically, the model outputs logits at intermediate layers to evaluate the effects of early output on task performance. Since this experiment is conducted on VQA tasks, the answer must depend on visual understanding. Before layer 18, accuracy is nearly zero, indicating that the model has not yet fused image information to generate valid answers. After layer 18, the accuracy rises notably, showing that visual information has been integrated and the model starts producing correct, image-grounded answers.

This trend is consistent with the observations in \Cref{sec:Image-Text Fusion Stage}, where performance stabilizes after layer 18, indicating that the model begins generating meaningful, visually grounded text once the fusion process is complete. Even when the output is produced earlier, such as at layer 18, the model can still provide partially correct answers, and the accuracy improves as the output layer moves deeper.

\section{Contrastive Attention: A Training-Free Method for Visual Refinement}


To investigate how the visual attention shifts of model during task understanding, we compare two representative layers: a pre-integrated layer and a post-integrated layer. Follow previous work \cite{zhang2025mllms}, the post-integrated layer is defined as the final layer, where task semantics and visual information have been fully integrated. In contrast, the pre-integrated layer corresponds to an earlier stage that captures the initial, task-agnostic perception of the image. By contrasting the attention between these two layers, we aim to reveal which image regions become increasingly aligned with task objectives as model processes the input.

In shallow layers, MLLMs typically exhibit broad, noisy attention over visually salient but task-irrelevant regions. To suppress such high-attention noise, we introduce a contrastive attention mechanism that computes the difference between the attention maps of the pre-integrated layer and the final (post-integrated) layer.  
To identify the pre-integrated layer, we compute the Hellinger distance between each layer’s attention map and that of the final layer, and select the one with the largest distance as the pre-integrated layer.
This choice allows the contrastive attention to capture the maximum representational shift between early perception and final understanding, providing richer information about how the model’s visual focus evolves throughout the reasoning process.

Besides, we explore multiple strategies to define the candidate set $\mathcal{C}$  for selecting the pre-integrated layer: (a) without any constraint, (b) constrained to shallow layers (layers before 16), (c) constrained to deep layers (layers after 17), and (d) constrained to fusion layers set $\mathcal{S}$ in \Cref{sec:Shallow Fusion Layers}. Details experiments are provided in \Cref{sec:Selection Strategy and Distribution of Pre-Integrated Layers}.

\begin{figure}[htbp]
    \centering
    \includegraphics[width=\linewidth]{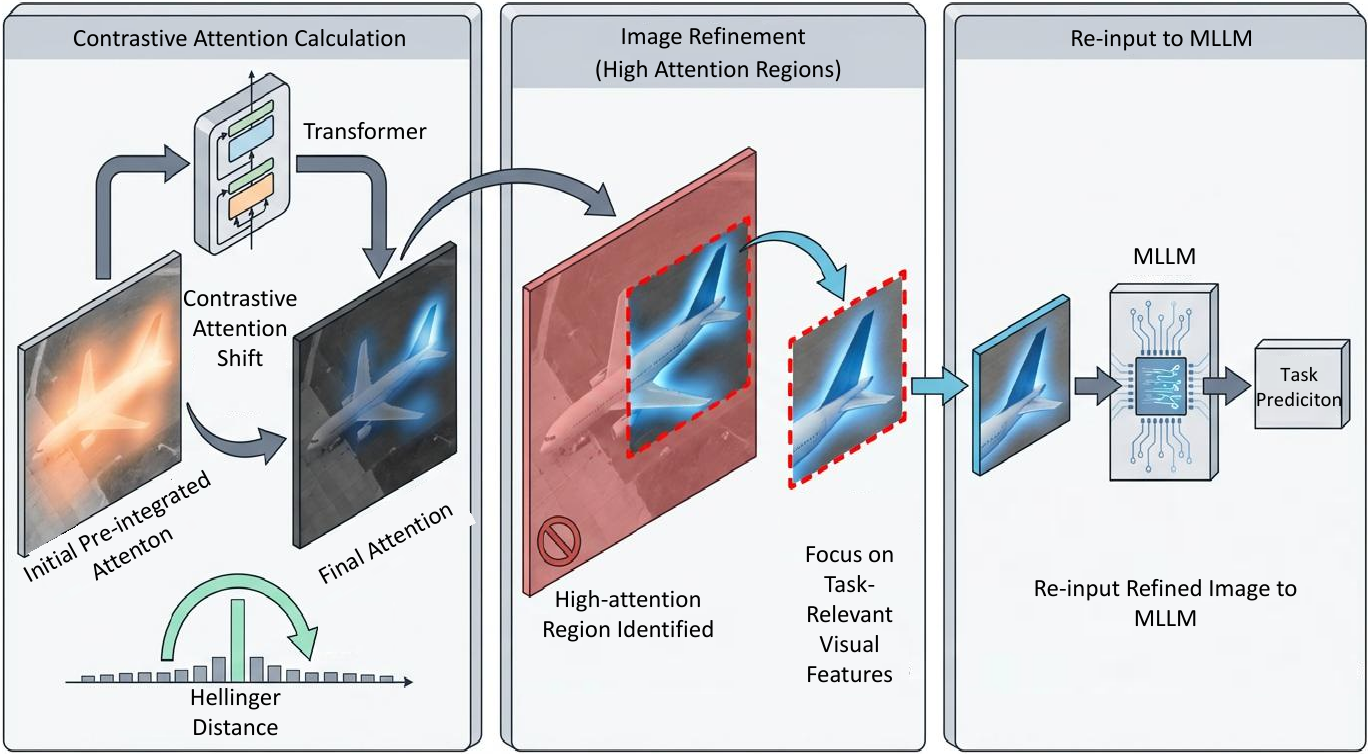}
    \caption{Mechanism of contrastive attention for the refined fusion-critical visual regions.}
    \label{fig:model}
\end{figure}

We formalize the selection process as follows: given a set of candidate attention matrices ${A^{(i)}}$, where each $A^{(i)} \in \mathbb{R}^{d \times d}$ represents the attention weights at layer $i$, we define the attention at the post-integrated layer (layer $k = 28$) as $A^{(k)}$.
To identify the attention matrix that deviates most significantly from $A^{(k)}$ in terms of distributional difference, we compute the Hellinger distance between each $A^{(i)}$ and the reference $A^{(k)}$, and select the one with the maximal distance from candidate layers $\mathcal{C}$:
    \begin{equation}
        i^* = \arg\max_{i \in \mathcal{C}} H\left(A^{(i)}, A^{(k)}\right)
        \label{eq:max_dis}
    \end{equation}
where $ H(P, Q) $ denotes the Hellinger distance between two probability distributions $ P $ and $ Q $, defined as:
\begin{equation}
    H(P, Q) = \frac{1}{\sqrt{2}} \sqrt{\sum_{j=1}^{d} \left( \sqrt{p_j} - \sqrt{q_j} \right)^2}
\end{equation}
where $d$ denotes the number of image tokens involved in the attention distribution.

\begin{table*}[htbp]
  \centering
  \caption{Effectiveness of contrastive attention compared to previous methods.}
  \renewcommand{\arraystretch}{.91}
  \resizebox{.95\linewidth}{!}{
    \begin{tabular}{clcccccccc}
    \toprule
          &       & GQA   & VQAv2 & OKVQA & VizWiz & TextVQA & DocVQA & MMBench & Average \\
    \midrule
    \multirow{6}[2]{*}{LLaVA\cite{liu2023llava}} & Base  & 66.03  & 74.06  & 51.59  & 57.85  & 57.21  & 24.57  & 63.08  & 56.34  \\
          & CD\cite{li2022CD} & 68.10  & 77.56  & 50.21  & 57.87  & 59.60  & 26.04  & 62.11  & 57.36  \\
          & DoLA\cite{chuang2023dola} & 66.83  & 76.11  & 52.84  & 58.45  & 56.11  & 26.03  & 63.31  & 57.10  \\
          & OPERA\cite{huang2024opera} & 68.56  & 76.42  & 52.99  & 58.61  & 58.72  & \textbf{27.45} & 64.04  & 58.11  \\
          & ViCrop\cite{zhang2025mllms} & 67.00  & 76.02  & 54.87  & 59.66  & 56.67  & 25.11  & 63.84  & 57.60  \\
          & Ours  & \textbf{69.40} & \textbf{77.59} & \textbf{55.42} & \textbf{60.35} & \textbf{59.86} & 26.98  & \textbf{64.49} & \textbf{59.16} \\
    \midrule
    \multirow{6}[2]{*}{LLaVA-Next\cite{liu2023improvedllava}} & Base  & 69.32  & 79.89  & 52.49  & 59.66  & 72.03  & 65.82  & 64.33  & 66.22  \\
          & CD\cite{li2022CD} & 70.59  & 78.47  & 50.11  & 58.97  & 71.66  & 66.43  & 63.10  & 65.62  \\
          & DoLA\cite{chuang2023dola} & 71.03  & 75.35  & 53.65  & 60.93  & 72.11  & 67.21  & 64.63  & 66.42  \\
          & OPERA\cite{huang2024opera} & 71.44  & 80.23  & 55.79  & 61.46  & 73.48  & 66.94  & 64.80  & 67.73  \\
          & ViCrop\cite{zhang2025mllms} & 70.20  & \textbf{81.40} & 54.10  & 61.30  & 73.49  & 65.68  & \textbf{65.60} & 67.40  \\
          & Ours  & \textbf{71.60} & 80.66  & \textbf{56.90} & \textbf{62.00} & \textbf{75.80} & \textbf{68.10} & 65.48  & \textbf{68.65} \\
    \midrule
    \multirow{6}[2]{*}{Qwen2.5VL\cite{qwen2.5-VL}} & Base  & 69.10  & 81.54  & 48.74  & 68.00  & 85.70  & 90.40  & 82.60  & 75.15  \\
          & CD\cite{li2022CD} & 70.89  & 83.02  & 48.44  & 69.02  & 86.28  & 91.81  & 81.33  & 75.83  \\
          & DoLA\cite{chuang2023dola} & 69.94  & 82.20  & 49.32  & 69.71  & 85.05  & 91.77  & 82.90  & 75.84  \\
          & OPERA\cite{huang2024opera} & 71.75  & 82.14  & 50.95  & 70.56  & 86.96  & 92.00  & 83.86  & 76.89  \\
          & ViCrop\cite{zhang2025mllms} & 71.12  & 84.10  & 51.84  & 71.13  & 86.89  & \textbf{92.39} & 83.60  & 77.29  \\
          & Ours  & \textbf{71.98} & \textbf{84.43} & \textbf{52.36} & \textbf{71.94} & \textbf{87.57} & 91.27  & \textbf{84.05} & \textbf{77.66} \\
    \bottomrule
    \end{tabular}%
}
  \label{tab:method_comparison}
\end{table*}

The contrastive attention is computed as the difference between the attention maps of a pre-integrated layer $ \mathbf{A}^{(i^*)} $ and the final layer $ \mathbf{A}^{(k)} $: $\text{IA} =  \mathbf{A}^{(k)} - \mathbf{A}^{(i^*)}$.
As illustrated in \Cref{fig:model}, we use this contrastive attention to guide image-level segmentation and re-input.
This difference highlights image regions where attention changes most significantly during multimodal fusion, indicating areas that the model gradually learns to associate with the task.
We then segment the original image based on this signal, selecting patches with high contrastive attention values, which correspond to the visual regions most informative for reasoning.
The segmented image $I_{\text{seg}}$ is then combined with the original image $I_{\text{orig}}$ and fed back into the MLLM for a second inference pass:
This re-input mechanism allows the model to explicitly revisit its own high-attention regions, reinforcing task-relevant visual cues while maintaining global context from the full image.

\subsection{Results Analysis}


We compare our method with representative training-free approaches:
(1) CD \cite{li2022CD} contrasts log-probabilities between an “expert” and an “amateur” LM to enhance fluency and factual consistency.
(2) DoLA \cite{chuang2023dola} contrasts logits between early and late layers to amplify deep factual knowledge and reduce hallucination.
(3) OPERA \cite{huang2024opera} penalizes over-trust in self-attention by reallocating attention weights.
(4) ViCrop \cite{zhang2025mllms} manipulates image inputs through masking and cropping to probe based on specific MLLM layer.
As shown in \Cref{tab:method_comparison}, our method consistently outperforms existing approaches. Compared with previous training-free decoding methods, our method achieves the highest average accuracy, indicating a stronger understanding of visual cues. 

Unlike CD and DoLA, which mainly contrast logits to enhance factual consistency, or OPERA, which adjusts decoding confidence, our approach directly optimizes the visual input through contrastive attention guided segmentation and re-input. Similar to ViCrop, our method follows a re-input paradigm that refines the visual evidence by reintroducing critical image regions. However, ViCrop requires manually determining the appropriate visual feature layer for each specific model, such as different layers for InstructBLIP and LLaVA-1.5, and tuning several model-dependent hyperparameters. The optimal cropping layer may also vary across datasets, limiting its general applicability. 
In contrast, we propose a dynamic layer-contrast selection strategy that automatically identifies the pre-integrated layer based on shallow-layer attention differences, providing a more stable and model-agnostic criterion for image segmentation. Furthermore, by leveraging our analysis of the fusion-layer mechanism, we introduce a fusion-based contrast selection strategy that constrains the pre-integrated layer choice and generalizes well across diverse architectures such as the LLaVA and Qwen series.

\subsection{Analysis of Pre-Integrated Layers}
\label{sec:Selection Strategy and Distribution of Pre-Integrated Layers}
Building on the results presented above, we further analyze the characteristics of the pre-integrated layers that serve as references for contrastive attention. This section examines how these layers are selected and their distribution.

\subsubsection{Selection Strategy}

\begin{figure}[htb]
    \centering
    \includegraphics[width=.95\linewidth]{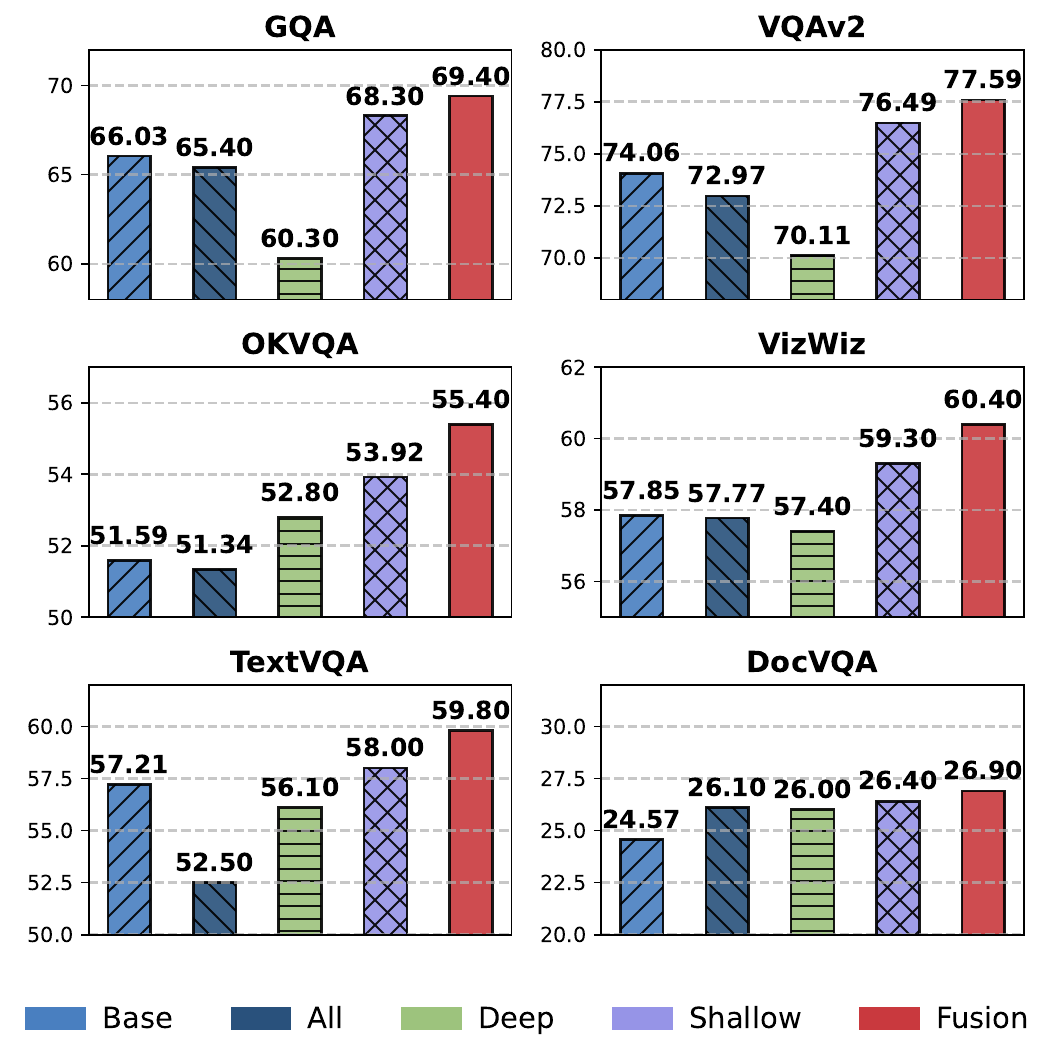}
    \caption{Layer selection strategies for pre-integrated layer.}
    \label{fig:selection_strategy}
\end{figure}

We systematically explored different strategies for selecting the pre-integrated layer, considering four selection scopes from:  
(a) \textit{all} layers (unconstrained setting),  
(b) \textit{deep} layers ,  
(c) \textit{shallow} layers, and  
(d) the set of \textit{fusion} layers identified in \Cref{sec:Shallow Fusion Layers}.
The pre-integrated layer refers to the stage where visual information is initially processed, before being influenced by textual context. Therefore, layers with already fused representations should be excluded. As shown in \Cref{fig:selection_strategy}, the \textit{deep} strategy performs poorly because these layers contain cross-modal information, which degrades the quality of contrastive attention. Similarly, the \textit{all} strategy includes deep layers, introducing noisy and biased attention.
In comparison, the \textit{shallow} strategy achieves better results. The best performance comes from the \textit{fusion} strategy, which limits selection to empirically identified fusion layers in \Cref{sec:Shallow Fusion Layers}.

\subsubsection{Layer Distribution}
\begin{figure}[htb]
    \centering
    \includegraphics[width=.95\linewidth]{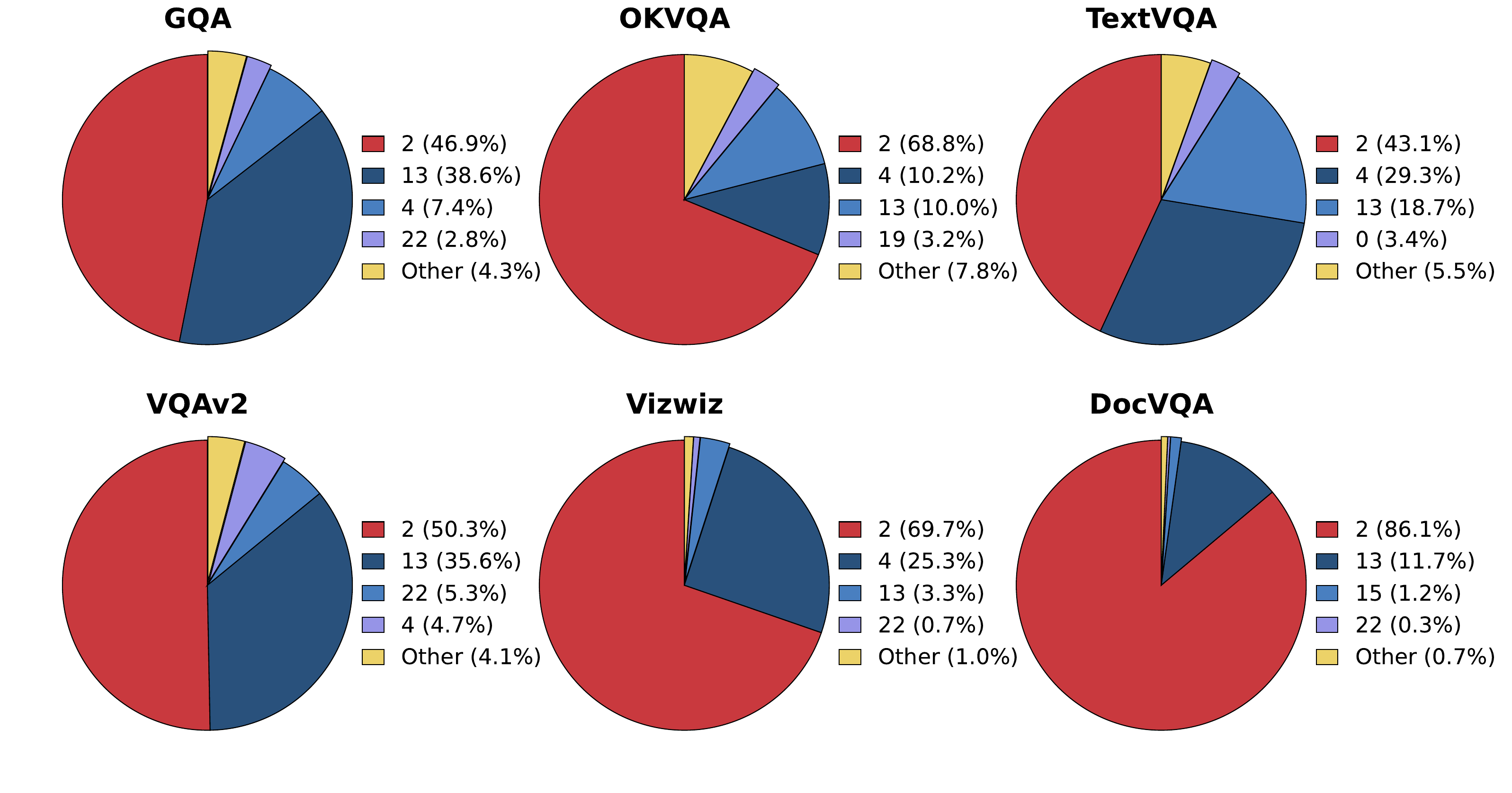}
    \caption{Distribution of selected pre-integrated layers across tasks
under unconstrained candidate setting.}
    \label{fig:jsd_layer_pies}
\end{figure}
As shown in Figure \ref{fig:jsd_layer_pies}, we further investigate the distribution of automatically selected pre-integrated layers when no constraints are imposed on the selection range. Specifically, we apply Equation~\eqref{eq:max_dis} to identify the layer that exhibits the largest attention divergence from the post-integrated layer. 
As illustrated, layer 2 is selected with overwhelmingly high frequency across all datasets, indicating a strong concentration around this early layer. For instance, in DocVQA, over 86\% of samples select layer 2 as the pre-integrated layer. 
\Cref{fig:jsd_layer_pies} shows that fusion layers $\mathcal{S}$ are favored as pre-integrated layers, reinforcing our claim that they are well-suited for contrastive attention.

\subsection{Inference Time Analysis}

\begin{figure}[htb]
    \centering
    \includegraphics[width=0.7\linewidth]{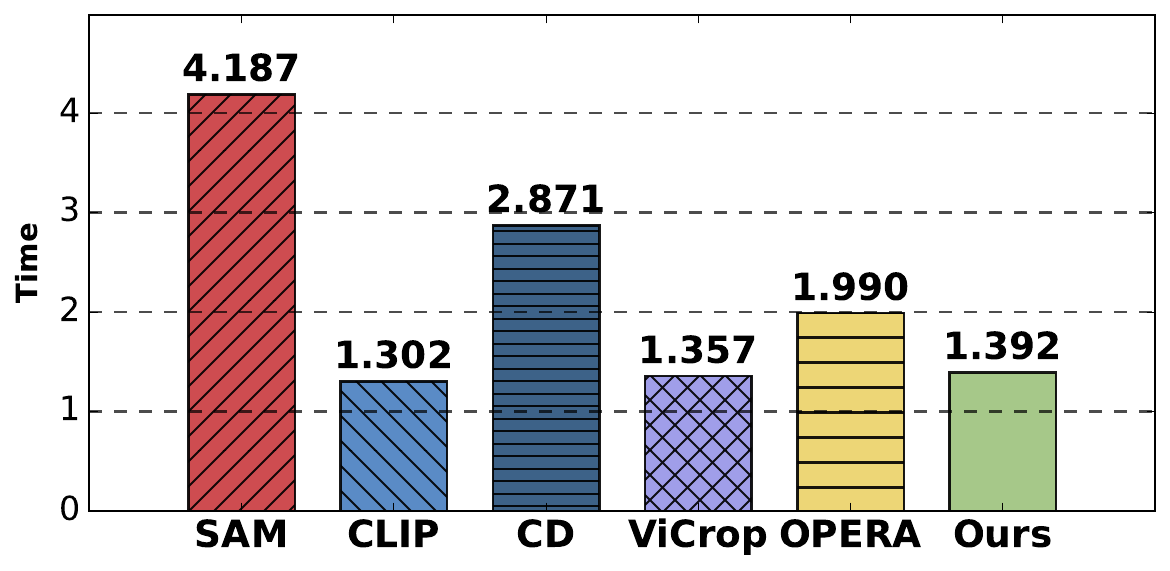}
    \caption{Inference time comparison on GQA (in seconds).}
    \label{fig:time}
\end{figure}
As shown in \Cref{fig:time}, we evaluate the inference time overhead of different training-free enhancement methods on the GQA dataset using an RTX A6000 GPU.
Although CLIP and ViCrop achieve slightly faster inference, their overall performance remains inferior to ours. In contrast, our method maintains a comparable inference time while achieving significantly better accuracy. The additional cost introduced by contrastive attention computation is minimal, indicating that our approach offers a favorable balance between efficiency and performance.




\section{Conclusion}
We systematically investigate how multimodal fusion emerges and evolves within MLLMs. Our analysis reveals that visual attention in MLLMs is often affected by persistent high-attention noise, which propagates across layers and distracts the model’s focus from task-relevant regions. To mitigate this issue, we introduce contrastive attention, which contrasts the attention between pre-integrated and final layers to capture text-guided attention shifts while filtering out semantically irrelevant activations. Through layer-wise masking and attention-distance analysis, we identify the key fusion layers that dominate vision–language integration and observe a consistent early-to-middle fusion pattern across models. Moreover, our findings highlight a “review” phenomenon in certain architectures, where visual signals are briefly reactivated at deep layers. 
Extensive experiments across multiple multimodal benchmarks demonstrate that our method outperforms previous methods.


\clearpage

\section*{Acknowledgements}
This work was supported by the New Generation Artificial Intelligence-National Science and Technology Major Project (2025ZD0123602),
the Science and Technology Innovation Program of Hunan Province under Grant 2024RC1048, and the National Key Laboratory Foundation Project under Grant 2024-KJWPDL-14.
This work was supported by the National Natural Science Foundation of China under Grant 62302144, 62401244, 72188101, 62307015, and 62437002.
    
{
    \small
    \bibliographystyle{ieeenat_fullname}
    \bibliography{main}
}


\end{document}